\documentclass{article}
\usepackage{spconf,amsmath,epsfig}
\usepackage{glossaries}
\usepackage{amsmath}
\usepackage{algorithmic}
\usepackage{graphicx}
\usepackage{textcomp}
\usepackage{xcolor}
\usepackage[font=footnotesize]{caption}
\usepackage{subcaption}
\usepackage{multirow}
\usepackage{adjustbox}
\usepackage{bm}
\usepackage{booktabs} % for better table rules
\usepackage{tabularx} % for better control over table width
\let\OLDthebibliography\thebibliography
\renewcommand\thebibliography[1]{
  \OLDthebibliography{#1}
  \setlength{\parskip}{0pt}
  \setlength{\itemsep}{0pt plus 0.3ex}
}
\newcommand{\mycustomsize}{\fontsize{7}{8}\selectfont}
\begin{document}\sloppy

% Example definitions.
% --------------------
\def\x{{\mathbf x}}
\def\L{{\cal L}}

% Title.
% ------
\title{FSDR: A Novel Deep Learning-based Feature Selection Algorithm for Pseudo Time-Series Data using Discrete Relaxation}
%
% Single address.
% ---------------
\name{Mohammad Rahman$^{1}$ \qquad Manzur Murshed $^{2}$ \qquad Shyh Wei Teng $^{1}$ \qquad Manoranjan Paul $^{3}$}
  
\address{$^{1}$ Federation University Australia, Australia \\
      $^{2}$ Deakin University, Australia \\
      $^{3}$ Charles Sturt University, Australia}

\maketitle

\newacronym{soc}{SOC}{Soil Organic Carbon}
\newacronym{rf}{RF}{Random Forests}
\newacronym{ann}{ANN}{Artificial Neural Network}
\newacronym{lucas}{LUCAS}{Land Use/Cover Area frame statistical Survey}

\newacronym{pca}{PCA}{Principal Component Analysis}
\newacronym{lle}{LLE}{Locally Linear Embedding}
\newacronym{lda}{LDA}{Linear Discriminant Analysis}
\newacronym{rfe}{RFE}{Recursive Feature Elimination}
\newacronym{sfs}{SFS}{Sequential Forward Selection}
\newacronym{sbs}{SBS}{Sequential Backward Selection}
\newacronym{rffi}{RFFI}{Random Forest Feature Importance}

\begin{abstract}
Conventional feature selection algorithms applied to Pseudo Time-Series (PTS) data, which consists of observations arranged in sequential order without adhering to a conventional temporal dimension, often exhibit impractical computational complexities with high dimensional data. To address this challenge, we introduce a Deep Learning (DL)-based feature selection algorithm: Feature Selection through Discrete Relaxation (FSDR), tailored for PTS data. Unlike the existing feature selection algorithms, FSDR learns the important features as model parameters using discrete relaxation, which refers to the process of approximating a discrete optimisation problem with a continuous one. FSDR is capable of accommodating a high number of feature dimensions, a capability beyond the reach of existing DL-based or traditional methods. Through testing on a hyperspectral dataset (i.e., a type of PTS data), our experimental results demonstrate that FSDR outperforms three commonly used feature selection algorithms, taking into account a balance among execution time, $R^2$, and $RMSE$.

\end{abstract}
\begin{keywords}Feature Selection, Band Selection, Discrete Relaxation, Continuous Approximation, Pseudo Time-Series Data, Gradient-based Search\end{keywords}
\section{Introduction}
\label{sec:intro}

%~\cite{stergioulas2022crop} 
%~\cite{sanchez2018bayesian}
%~\cite{DAGLIATI2020101930}
The application of pseudo time-series (PTS) data ~\cite{peeling2007making}, characterized by sequential patterns analogous to traditional time-series data but with the absence of a conventional temporal dimension, has garnered substantial interest across diverse fields, including remote sensing, bioinformatics, and medicine. For example, in hyperspectral imaging, the sequence of bands within each pixel forms a pseudo time-series, capturing variations in spectral signatures. Effective analyses of these variations are crucial for estimating various physical and chemical properties from these data. In scientific research, obtaining a low-dimensional representation of high-dimensional hyperspectral data is important for purposes such as enhancing computational efficiency and identifying the most relevant features for specific tasks. This dimensionality reduction is mainly achieved through two methods: Feature Extraction (FE) and Feature Selection (FS) ~\cite{zebari2020comprehensive}.

%Hyperspectral devices feature hundreds or even thousands of spectral bands, rendering hyperspectral data inherently high-dimensional.

%
%~\cite{pearson1901liii}
%~\cite{chen2011locally}
% ~\cite{fisher1936use}
FE algorithms transform the original data into a reduced-dimensional representation, capturing the most relevant information. \acrfull{pca}, \acrfull{lle}, and \acrfull{lda} are examples of such algorithms. With the advent of deep learning (DL), numerous DL-based methodologies have been proposed for feature extraction in recent years ~\cite{rasti2020feature}. However, dimensionality reduction with FE algorithms comes at the cost of interpretability ~\cite{li2020toward}, leading to the loss of a direct understanding of individual features in the process. Therefore, when interpretability is desirable, FE algorithms may not be the best choices ~\cite{shah2019feature}.

%verma2020feature
To maintain interpretability while reducing data dimensionality, FS algorithms are typically employed. These algorithms involve finding a subset of relevant features that can be selected depending on defined criteria ~\cite{zebari2020comprehensive}. Selecting a specified target number of optimal features from the original feature set for a particular machine learning task can be formulated as a combinatorial optimisation problem ~\cite{WANG201621}, and hence many problems related to feature selection are NP-Hard ~\cite{937619}. Therefore, traditional FS algorithms, which rely on heuristic searching, are unable to provide any guarantee of optimality ~\cite{sun2007iterative}. All these algorithms have their own pros and cons based on the heuristics they use.

%~\cite{dhumkekar2022performance}
%~\cite{thomas1991elements}
%~\cite{li2005analysis}
%~\cite{ververidis2005sequential}
%~\cite{tibshirani1996regression}
%~\cite{hoerl1970ridge}
The most common FS algorithms can be classified into: (a) Filter, (b) Wrapper, and (c) Embedded methods ~\cite{zebari2020comprehensive}. Filter methods, e.g., Variance Threshold (VT) and Mutual Information (MI), select features based on certain statistical measures. While this makes them computationally efficient and independent of specific algorithms, they are unable to capture feature interactions or dependencies, as the assessment of features is performed independently. Wrapper methods, e.g., Recursive Feature Elimination (RFE) and Sequential Forward Selection (SFS), address this limitation by evaluating the performance of different combinations of features, albeit at a higher computational cost. The computational issue can be mitigated with Embedded methods, e.g., Least Absolute Shrinkage and Selection Operator (LASSO) and Ridge, where feature selection is integrated into the model training process for the specific task at hand. In recent years, a number of DL-based FS algorithms have been proposed ~\cite{bobadilla2021deep}, ~\cite{zou2015deep}, ~\cite{liu2021automated}. However, none of the existing FS algorithms are tailored for PTS data. Therefore, they cannot reduce time complexity by exploring through feature dimension with gradient-based search.

A major challenge in finding an optimal subset of features by exploring through the feature dimension with gradient-based search is that the gradient-based search is originally designed for continuous optimisation, whereas indices of features, which represent the positions of the features in the feature dimension, are discrete integers. A common solution to this issue is applying discrete relaxation, which involves transforming a discrete problem into a continuous one by allowing the variables to take on fractional values in a continuous domain ~\cite{hancock1990discrete}. By transforming discrete PTS data into continuous functions using discrete relaxation, we can allow the target indices to take on non-integer values, thereby transforming the problem into continuous optimisation, which can be tackled by gradient-based search. In the end, we can derive the target set by rounding each target index to the nearest integer.

Sheth et al.~\cite{sheth2020differentiable} demonstrated an effective way of utilizing discrete relaxation in feature selection. Their proposed method aims to achieve a set \(s \in \{0, 1\}^D\) through gradient-based search, where \(s_i = 1\) indicates that the \(i\)-th feature is present, while \(s_i = 0\) indicates its absence in the final set of selected features. They obtain such a set by optimizing a cost function that is inspired by a function proposed by Kong et al.~\cite{kong2018estimating}, which represents the residual variance estimate for a subset of features. In order to perform gradient-based search between discrete integers 0 and 1, they relaxed the discrete search domain to a continuous domain via a sigmoid transformation. However, in the context of PTS data, their method shares the same limitation with the aforementioned DL-based FS algorithms of not leveraging the sequential pattern in the data. Furthermore, they perform a gradient-based search over a relaxed continuous domain (ranging from 0 to 1) for all the original features, resulting in a time complexity associated with the dimensionality of the original features. We seek a solution where the time complexity depends on the number of targeted selected features but remains robust against the high dimensionality of the original features.

To disassociate time complexity from the original feature dimension, we frame the scenario of feature selection as a supervised machine learning problem to be addressed with a DL-based model that takes high-dimensional PTS data as input and the corresponding response variable as output. During the training process, only a subset of the original feature indices, matching the specified target size, is utilized, ensuring that the computational complexity depends on the specified target size, not the original feature size. It also indicates that the number of training samples required to train the algorithm does not depend on the original feature size, as the number of model parameters is solely influenced by the target size. This is a noteworthy trait as DL-based models are known to generally require a higher number of training samples compared to other types of algorithms ~\cite{janiesch2021machine}. Through the training process, the subset is updated to represent the set of important features for the supervised machine learning task in concern.

In this research, we propose a DL-based FS algorithm named Feature Selection through Discrete Relaxation (FSDR). FSDR effectively and efficiently selects important features in PTS data, addressing the limitations of traditional algorithms. Our primary contributions lie in the facts that: (a) to the best of our knowledge, FSDR is the first FS algorithm where target features are achieved by tuning learnable parameters across the feature dimension through gradient-based search; (b) FSDR is capable of accommodating a high number of original features without significantly affecting execution time, in contrast to existing FS algorithms; (c) FSDR is tolerant of limited training samples despite being a DL-based model.

The rest of the paper is structured as follows: Section 2 outlines the methodology used in the study. Section 3 presents our results in addition to our observations on them. Finally, Section 4 concludes the research and highlights the future directions of this research.

\section{Methodology}
\subsection{Datasets}

% ~\cite{WANG2022105241}
The Land Use/Cover Area frame statistical Survey (LUCAS) dataset, composed of 21,782 topsoil samples. It provides absorbance measurements of 4,200 spectral bands in addition to a set of soil properties. In this study, our objective is to develop an FS algorithm capable of identifying subsets of varying sizes that are important to estimate Soil Organic Carbon (SOC), which is available as ground truth in the LUCAS dataset. For improved prediction, the absorbance values have been transformed into reflectance values.

To assess the algorithm's performance under conditions of high dimensionality compared to low dimensionality and to evaluate its efficacy in scenarios with limited training samples, three different datasets are employed: (a) original, consisting of all available data samples and features; (b) downsampled, where all data points are retained but with a reduced feature size of 66 achieved through Discrete Wavelet Transform (DWT) downsampling; and (c) truncated, comprising a randomly sampled subset of 871 data points from the original dataset while retaining all available features.

It is noteworthy that within the realm of hyperspectral data, feature selection is more commonly addressed as band selection when the focus is solely on spectral bands. The discourse on band selection algorithms typically adopts a distinct categorization ~\cite{sun2019hyperspectral}, with certain algorithms being explicitly designed for hyperspectral data. However, to uphold a broad perspective on general PTS data in this study, we will refrain from delving into the specifics of band selection algorithms.

\subsection{Experimental Design and Objectives}
%= \{(\mathbf{X}_1, y_1), (\mathbf{X}_2, y_2), \ldots, (\mathbf{X}_N, y_N)\}
Let \( \mathcal{D}_{N \times D} \) represent a dataset with \(N\) samples and \(D\) features, where \( \mathbf{X}_i \) is the \(i\)-th data sample, and \( y_i \) is the corresponding response variable, SOC. The set of original feature indices is denoted as \( F_D = [1, 2, 3, \ldots, D] \). To select a maximum of \(t\) features (\(t \leq D\)) important for estimating SOC, our objective is to derive \(F_{t'}\), where \(F_{t'} \subseteq F_D\), \(|F_{t'}| = {t'}\), and \(t' \leq t\).

For the three aforementioned datasets, we aim to find the target feature set using FSDR for target sizes 2, 5, 10, 15, and 20. To compare FSDR with existing FS algorithms, we replicate this entire experimental setup using three established traditional FS algorithms, each representative of a major group: (a) Filter method - MI, (b) Wrapper method - SFS, (c) Embedded method - LASSO. To ensure a fair and unbiased comparison of the algorithms, we adopt a uniform benchmarking approach. The efficacy of the feature set selected by each algorithm is evaluated using a standardized DL-based regression model with two hidden layers consisting of 15 and 10 nodes, respectively. For each combination of algorithm and target size, the dataset has been divided into 90\% for training and 10\% for testing. We monitor the time required to train the FS model in each case in seconds. In addition, after an FS model provides the selected features for a combination, with the training data, we train the regression model using the features selected, and track the corresponding $R^2$ and root mean square error ($RMSE$) for the test data.

To further validate the effectiveness of FSDR, we independently assess the predictive performance of two feature sets: (1) the initial target set before the commencement of FSDR training, and (2) the final updated set after the training process.

It is worth mentioning that we are limiting the experiments to a maximum of 20 target features because LUCAS exhibits a high degree of multicollinearity. Consequently, using a number of features beyond this range does not yield significantly higher accuracy unless employing more advanced algorithms that require all the features ~\cite{tsakiridis2020simultaneous}.

\subsection{Motivation}

\begin{figure}[h]
\centering
\includegraphics[width=0.5\textwidth]{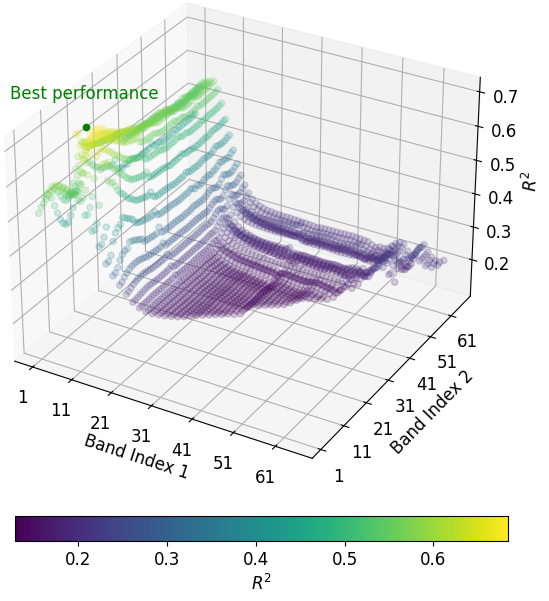}
\caption{Model performance ($R^2$) with all combinations of two indices from 66 bands in the downsampled dataset. It is clear that the model performance smoothly changes with small changes in band indices. }
\label{fig:motivation}
\end{figure}

For a model that predicts SOC from multiple bands, in order to visualize how the performance of the model is affected with small changes in the band indices, we take all possible combinations of two bands from the 66 bands available in the downsampled dataset. To avoid duplications, we only considered cases where the second band is higher than the first. For each combination, we train a DL model to predict SOC. The model performance ($R^2$) for each combination is illustrated in Fig. \ref{fig:motivation}. As evident from the figure, with a small change in an index, the model performance changes smoothly. It inspires us to develop an algorithm where the target indices are learned through a neural network based on the Mean Squared Error (MSE) in model performance for small adjustments in the band indices. One challenge in the process is that the band indices are discrete integers between 1 and 66, and undefined for non-integer values of the bands, represented by the gaps in the Fig. \ref{fig:motivation}. To address this, we need to employ discrete relaxation. By utilizing an interpolation method, we must ensure that band reflectances are defined for any non-integer value between consecutive indices, transforming the problem into a continuous optimization task, which can be tackled through a gradient-based search within a DL-based architecture.

\subsection{Proposed Algorithm}
\begin{figure}[h]
  \centering
  
  \begin{subfigure}{0.5\textwidth}
    \centering
    \includegraphics[width=\linewidth]{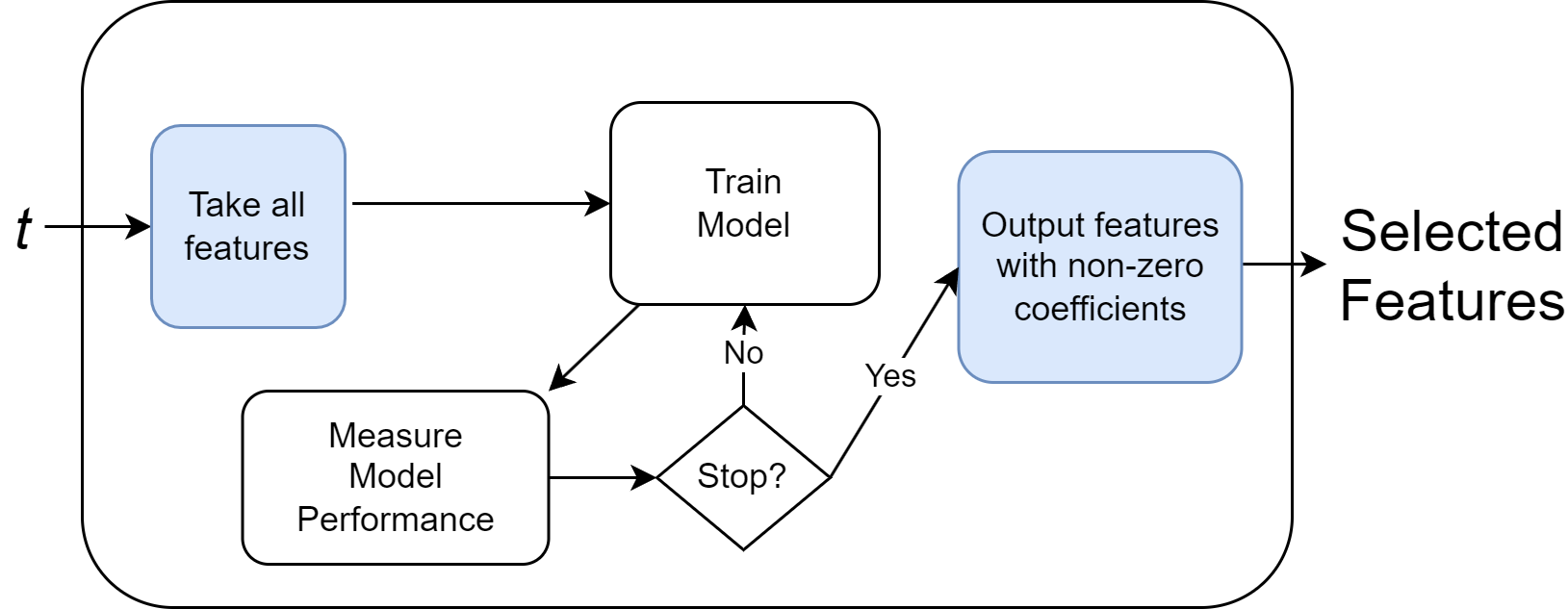}
    \caption{LASSO}
    \label{fig:lasso}
  \end{subfigure}

  \begin{subfigure}{0.5\textwidth}
    \centering
    \includegraphics[width=\linewidth]{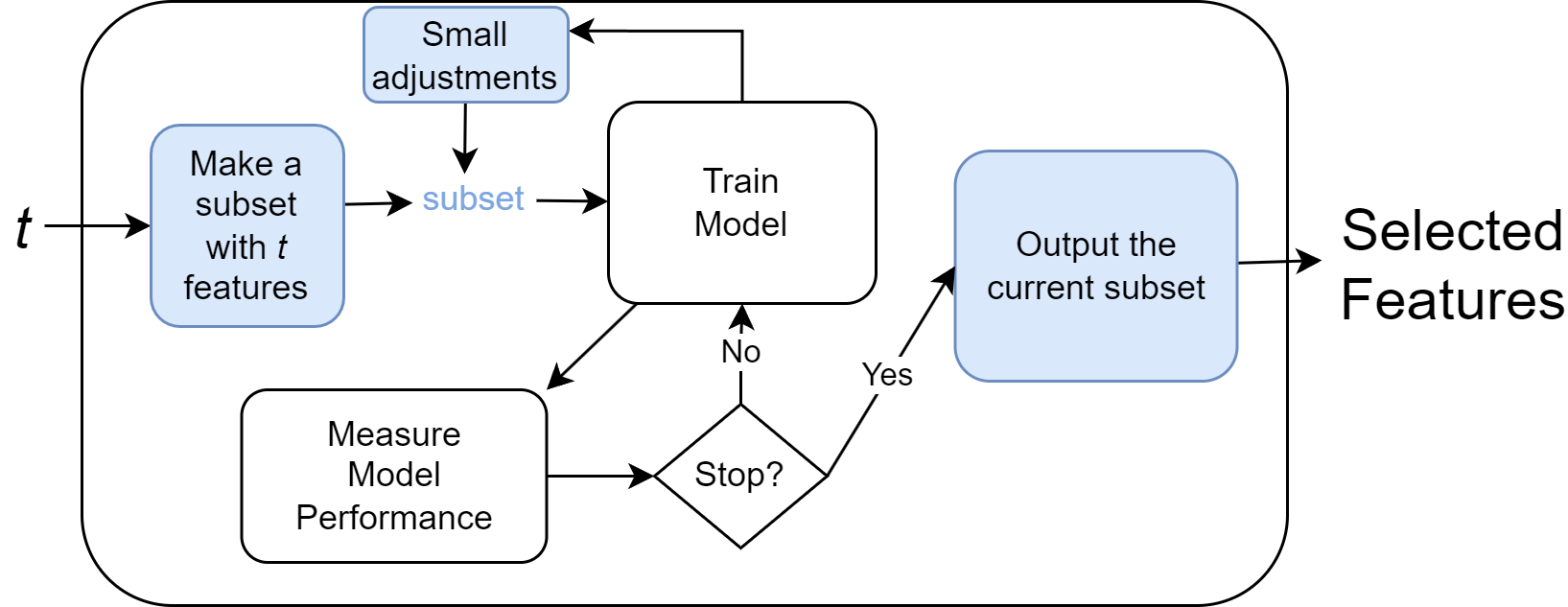}
    \caption{FSDR}
    \label{fig:fsdr}
  \end{subfigure}
  
  \caption{Contrasting the high-level architecture of (b) FSDR with (a) LASSO, highlighting the major differences. LASSO trains the model with all the available features. On the other hand, FSDR takes a subset with the given target number ($t$) of features, trains the model, updates the subset with nearby features of the current features in the subset based on their impact on the model performance.}
  \label{fig:arch}
\end{figure}
FSDR focuses on updating the target feature subset for better prediction of SOC. Due to the sequential pattern across the feature dimension, the indices in the subset can be updated based on the impact of small adjustments, thereby eliminating the need to assess model performance for arbitrary combinations of features and preventing combinatorial explosion. To perform the update through gradient-based search, the indices are relaxed to a continuous domain via discrete relaxation. Upon successful training, we not only obtain a supervised learning model but also the desired subset of features, classifying FSDR as a type of embedded method. A typical embedded FS algorithm optimizes both model parameters and relevant feature subsets simultaneously to improve predictive performance. For example, LASSO begins training the model with all available features, aiming to set the coefficients of irrelevant features to 0. At the end of the training, the resulting subset is determined by selecting the top $t$ features with non-zero coefficients in the model as depicted in Fig. \ref{fig:lasso}. The size of the final set, denoted as ${t'} = |F_{t'}|$, may be less than $t$ if there are fewer than $t$ features available with non-zero coefficients.

\begin{figure}[h]
\centering
\includegraphics[width=0.5\textwidth]{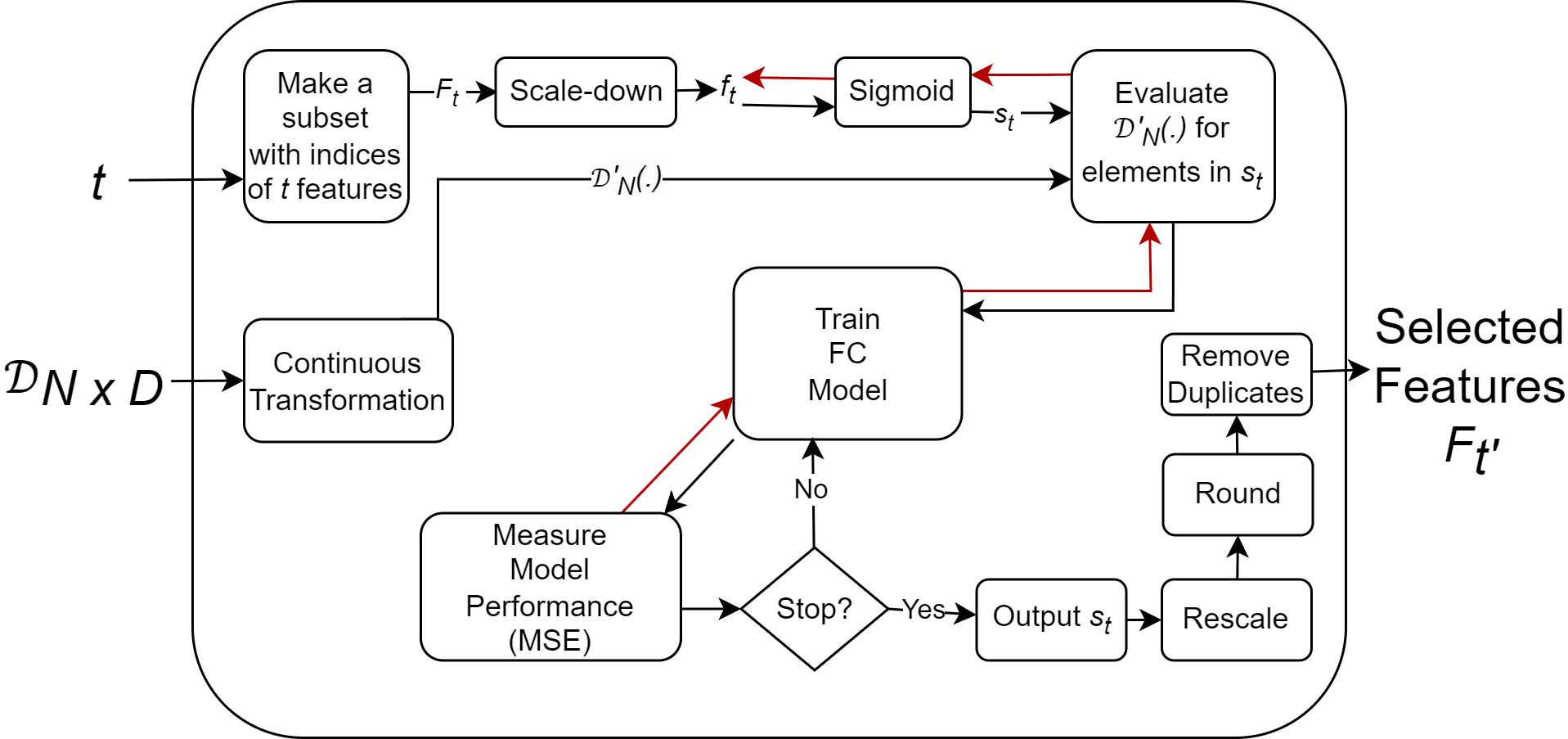}
\caption{Detailed FSDR Architecture. Training starts with an initial subset $F_t$ containing indices of $t$ features, which is transformed into $s_t$ through a sequence of scale-down and sigmoid transformations. Dataset $\mathcal{D}_{N \times D}$ is transformed into $N$ continuous functions $\mathcal{D'}_{N}(.)$. For each element in $s_t$, $\mathcal{D'}_{N}(.)$ is evaluated and passed to an FC network, and $f_t$ is updated through backpropagation. The final set $F_{t'}$ is obtained from $s_t$ by rescaling, rounding each element to the nearest integer, thereby transforming the elements back to the representation of actual indices, and removing duplicates.
}
\label{fig:fsdr2}
\end{figure}

The most important difference in FSDR from LASSO and other embedded FS algorithms is that FSDR initiates training with a subset of $t$ elements, $F_t$ from the original features, $F_D$, as shown in Fig. \ref{fig:fsdr}, rather than using all $D$ features. Applying the gradient-descent algorithm, subset elements are updated with nearby features based on their impact on model performance, with the goal of representing the desired subset by the end of training.

Fig. \ref{fig:fsdr2} illustrates a more detailed view of the FSDR architecture. The subset of $t$ features $F_t$ is initialized with indices linearly spaced, covering $t$ indices between 1 and $D$, with the aim that it is gradually updated through training to represent the desired target subset. In order to process actual indices values ranging from 1 to $D$ through neural networks without causing exploding gradient issues, we first transform the set $F_t$ to $f_t$ by scaling down its elements. We then transform $f_t$ to $s_t$ through a sigmoid transformation, ensuring a consistent mapping between $F_t$ and $s_t$. An element of 0 in $s_t$ indicates the first actual feature, while an element of 1 in $s_t$ corresponds to the $D$-th actual feature.

\begin{figure}[h]
\centering
\includegraphics[width=0.5\textwidth]{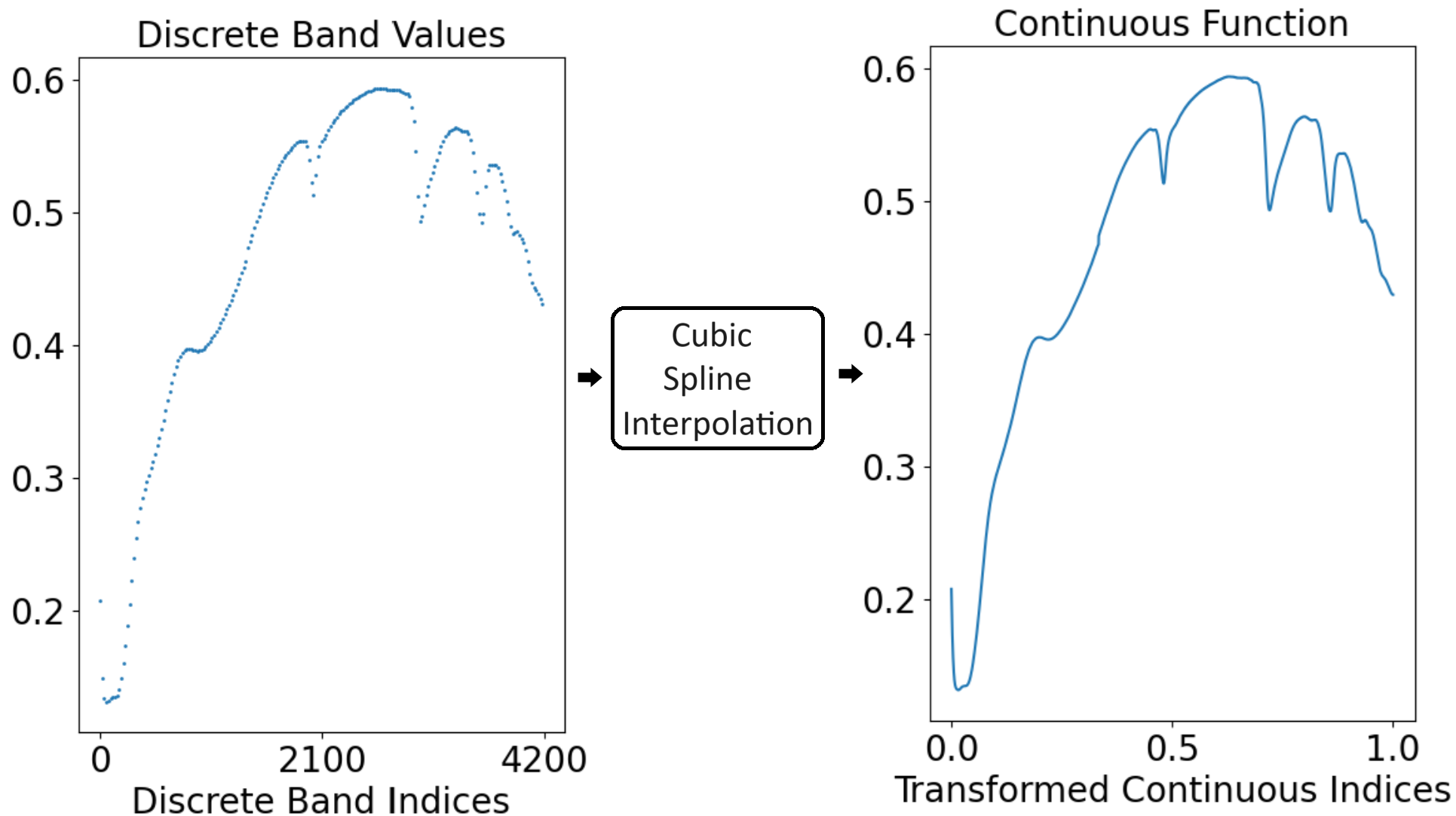}
\caption{All the discrete features (band reflectances) for each sample are transformed into a continuous function with a domain ranging from 0 to 1 using Cubic Spline Interpolation.}
\label{fig:cubic}
\end{figure}

To ensure that the band reflectances can be evaluated for the indices represented by the non-integer values in \(s_t\), ranging between 0 and 1, through the application of discrete relaxation, the non-negative integer domain of the feature indices is transformed into a continuous domain. For each data sample in \(\mathcal{D}_{N \times D}\), we transform the \(D\) discrete features, representing the reflectance of \(D\) bands, into a continuous function with the domain between 0 and 1 using an interpolation method, as illustrated in Fig. \ref{fig:cubic}. This interpolation method yields a function defined for all indices, including non-integer values between each consecutive pair of indices. We employed Cubic Spline Interpolation as interpolation method, as this technique ensures the generation of smooth interpolating curves and guarantees first derivative continuity at each interpolation point. Preserving such continuity is particularly advantageous in the context of gradient calculation during the backpropagation phase of the neural network training process.

The set of functions that are procured through the interpolation method is denoted by \( \mathcal{D'}_{N}(.) \). For each element in \(s_t\), \( \mathcal{D'}_{N}(.) \) is evaluated and passed to a training model. The training model in this case is a Fully Connected (FC) neural network, with two hidden layers of 15 and 10 nodes, respectively, which is trained by optimizing an MSE loss function. Throughout the training process, not only are the network parameters of FC updated, but also $f_t$ undergoes updates through backpropagation, as indicated by the red arrows in Fig. \ref{fig:fsdr2}. After the training is finished, the values of $s_t$ are rescaled and rounded to the nearest integer value to transform them back into the domain of the actual indices. Finally, we achieve the desired set $F_{t'}$ by removing the duplicate elements from the rounded values.

\section{Results and Discussions}
\label{sec-res}
\subsection{Results for original, downsampled, and truncated datasets}

\begin{table}[ht]
\centering
\mycustomsize
\caption{Execution time ($time$) in seconds, $R^2$, and $RMSE$ of the algorithms for three datasets and five target sizes ($t$). The best (bold) and the second best (blue and italicized) scores for each combination of the dataset and target size are highlighted. While $R^2$ and $RMSE$ for FSDR are comparable to those of SFS, as the target size and original feature size increase, the execution time for FSDR does not increase as sharply as for SFS.}
\label{result-all}

\begin{tabular}
{|l|l|r|r|r|r|r|}\hline Metric & Dataset & t  & MI & SFS & LASSO & FSDR \\ \hline
\multirow{15}{*}{$time$ (sec)} 
	 & \multirow{5}{*}{original}
 	  & 2  & 479.78  & \textit{\textcolor{blue}{56.50}}  & 153.82  & \textbf{32.39} \\
 &	  & 5  & 447.06  & 200.98  & \textit{\textcolor{blue}{153.97}}  & \textbf{45.37} \\
 &	  & 10  & 411.33  & 637.72  & \textit{\textcolor{blue}{153.04}}  & \textbf{66.36} \\
 &	  & 15  & 411.76  & 1208.30  & \textit{\textcolor{blue}{152.63}}  & \textbf{89.57} \\
 &	  & 20  & 411.71  & 2071.68  & \textit{\textcolor{blue}{153.62}}  & \textbf{110.42} \\
\cline{2-7}
	 & \multirow{5}{*}{downsampled}
 	  & 2  & 6.43  & \textit{\textcolor{blue}{0.98}}  & \textbf{0.15}  & 26.82 \\
 &	  & 5  & 6.56  & \textit{\textcolor{blue}{3.12}}  & \textbf{0.15}  & 36.95 \\
 &	  & 10  & \textit{\textcolor{blue}{6.47}}  & 8.19  & \textbf{0.15}  & 52.59 \\
 &	  & 15  & \textit{\textcolor{blue}{6.54}}  & 14.78  & \textbf{0.14}  & 68.10 \\
 &	  & 20  & \textit{\textcolor{blue}{6.45}}  & 22.65  & \textbf{0.15}  & 86.25 \\
\cline{2-7}
	 & \multirow{5}{*}{truncated}
 	  & 2  & 12.76  & 34.92  & \textbf{1.07}  & \textit{\textcolor{blue}{10.79}} \\
 &	  & 5  & \textit{\textcolor{blue}{12.65}}  & 88.12  & \textbf{1.11}  & 14.57 \\
 &	  & 10  & \textit{\textcolor{blue}{12.74}}  & 188.66  & \textbf{0.99}  & 19.35 \\
 &	  & 15  & \textit{\textcolor{blue}{12.84}}  & 299.94  & \textbf{0.97}  & 24.52 \\
 &	  & 20  & \textit{\textcolor{blue}{12.83}}  & 437.30  & \textbf{1.10}  & 30.15 \\
\hline
\multirow{15}{*}{$R^2$} 
	 & \multirow{5}{*}{original}
 	  & 2  & 0.51  & \textbf{0.69}  & \textit{\textcolor{blue}{0.52}}  & 0.25 \\
 &	  & 5  & 0.51  & \textbf{0.83}  & 0.53  & \textbf{0.83} \\
 &	  & 10  & 0.61  & \textbf{0.86}  & 0.53  & \textit{\textcolor{blue}{0.84}} \\
 &	  & 15  & 0.61  & \textbf{0.87}  & 0.53  & \textit{\textcolor{blue}{0.83}} \\
 &	  & 20  & 0.54  & \textbf{0.87}  & 0.78  & \textit{\textcolor{blue}{0.84}} \\
\cline{2-7}
	 & \multirow{5}{*}{downsampled}
 	  & 2  & 0.53  & \textbf{0.69}  & 0.52  & \textit{\textcolor{blue}{0.55}} \\
 &	  & 5  & 0.69  & \textbf{0.82}  & 0.78  & \textit{\textcolor{blue}{0.80}} \\
 &	  & 10  & 0.72  & \textbf{0.85}  & 0.79  & \textit{\textcolor{blue}{0.83}} \\
 &	  & 15  & 0.72  & \textbf{0.87}  & 0.80  & \textit{\textcolor{blue}{0.82}} \\
 &	  & 20  & 0.79  & \textbf{0.86}  & 0.81  & \textit{\textcolor{blue}{0.82}} \\
\cline{2-7}
	 & \multirow{5}{*}{truncated}
 	  & 2  & 0.42  & \textbf{0.70}  & 0.07  & \textit{\textcolor{blue}{0.47}} \\
 &	  & 5  & 0.42  & \textbf{0.89}  & 0.82  & \textit{\textcolor{blue}{0.87}} \\
 &	  & 10  & 0.41  & \textit{\textcolor{blue}{0.90}}  & 0.82  & \textbf{0.91} \\
 &	  & 15  & 0.42  & \textbf{0.91}  & 0.90  & \textbf{0.91} \\
 &	  & 20  & 0.43  & \textit{\textcolor{blue}{0.90}}  & \textit{\textcolor{blue}{0.90}}  & \textbf{0.91} \\
\hline
\multirow{15}{*}{$RMSE$} 
	 & \multirow{5}{*}{original}
 	  & 2  & \textit{\textcolor{blue}{0.10}}  & \textbf{0.08}  & \textit{\textcolor{blue}{0.10}}  & 0.13 \\
 &	  & 5  & 0.10  & \textbf{0.06}  & 0.10  & \textbf{0.06} \\
 &	  & 10  & 0.09  & \textbf{0.05}  & 0.10  & \textit{\textcolor{blue}{0.06}} \\
 &	  & 15  & 0.09  & \textbf{0.05}  & 0.10  & \textit{\textcolor{blue}{0.06}} \\
 &	  & 20  & 0.10  & \textbf{0.05}  & 0.07  & \textit{\textcolor{blue}{0.06}} \\
\cline{2-7}
	 & \multirow{5}{*}{downsampled}
 	  & 2  & 0.10  & \textbf{0.08}  & 0.10  & \textit{\textcolor{blue}{0.09}} \\
 &	  & 5  & 0.08  & \textbf{0.06}  & 0.07  & \textbf{0.06} \\
 &	  & 10  & 0.07  & \textbf{0.06}  & \textbf{0.06}  & \textbf{0.06} \\
 &	  & 15  & 0.07  & \textbf{0.05}  & \textit{\textcolor{blue}{0.06}}  & \textit{\textcolor{blue}{0.06}} \\
 &	  & 20  & \textit{\textcolor{blue}{0.06}}  & \textbf{0.05}  & \textit{\textcolor{blue}{0.06}}  & \textit{\textcolor{blue}{0.06}} \\
\cline{2-7}
	 & \multirow{5}{*}{truncated}
 	  & 2  & \textit{\textcolor{blue}{0.14}}  & \textbf{0.10}  & 0.18  & \textit{\textcolor{blue}{0.14}} \\
 &	  & 5  & 0.14  & \textbf{0.06}  & 0.08  & \textit{\textcolor{blue}{0.07}} \\
 &	  & 10  & 0.14  & \textbf{0.06}  & 0.08  & \textbf{0.06} \\
 &	  & 15  & 0.14  & \textit{\textcolor{blue}{0.06}}  & \textit{\textcolor{blue}{0.06}}  & \textbf{0.05} \\
 &	  & 20  & 0.14  & \textbf{0.06}  & \textbf{0.06}  & \textbf{0.06} \\
\hline

\end{tabular}

\end{table}

Table \ref{result-all} illustrates the execution time in seconds, $R^2$, and $RMSE$ of the algorithms for different target sizes and datasets. While SFS seems to slightly outperform FSDR in terms of $R^2$ and $RMSE$, as indicated in Table \ref{result-all}, the effectiveness of FSDR becomes apparent when considering the increase in execution time with a larger original feature dimension. Although SFS is efficient with a smaller number of original features, such as in the case of the downsampled dataset with 66 features, it demands a considerable amount of time for a larger number of original features, as observed in the case of the original dataset with 4,200 features. This is due to the substantial increase in the number of feature combinations to assess with an expansion in the target size. On the other hand, in the case of FSDR, it is noteworthy that the original feature size does not significantly influence the execution time. An increase in the original feature size incurs additional time only for the construction and evaluation of a more detailed continuous function. However, this process can be significantly optimized by applying simpler interpolation methods than Cubic Spline Interpolation.

Moreover, as the target size increases with SFS, there is a notable rise in the execution time due to the influence of target size on the number of feature combinations to assess. In contrast, in FSDR, the increase in target size results in the addition of new weights only at the first layer of the neural network. These weights accommodate the learnable parameters representing the additional target indices. Therefore, the increased execution time remains manageable with an increase in the target size.

Since the number of model parameters in the case of FSDR depends on the target size, which is generally lower than the original feature size, a modest number of training samples are sufficient to train FSDR, as depicted in the case of truncated dataset, despite it being a DL-based model. Also, it is noteworthy that the execution time for FSDR significantly decreases with fewer training samples. Despite maintaining the same number of training epochs, a reduction in training samples results in fewer continuous functions to construct and evaluate, leading to decreased processing time.

While LASSO and MI are highly time-efficient when either the training sample size or the original feature size is low, they are significantly slower with a high training sample size and original feature size. In terms of $R^2$ and $RMSE$, FSDR consistently outperforms LASSO by a moderate margin, and MI by a high margin.

\subsection{Results for initialized target set compared to trained target set}

For the particular case of the original dataset with target size five, we obtain the initialized target set and the final target set after the training is finished. The effectiveness of both the initialized and final target sets is evaluated using a separate untrained DL-based model. The $R^2$ values for the initialized and the final target sets are 0.60 and 0.72, respectively. Similar enhancements are observed in all other cases involving FSDR. In the case of the original dataset with target size 5, initially, the set is populated with 5 linearly spaced indices ranging from 1 to 4,200 (696, 1,390, 2,082, 2,774, and 3,466). Throughout the training process, the set undergoes iterative updates to enhance its predictive capability for SOC estimation, as reflected in the $R^2$ values. This affirms that the efficacy of FSDR extends beyond the FC layer in the architecture, and the learnable parameters representing the target indices are indeed updated during training to improve the set's ability to predict SOC.

\section{Conclusion}
We demonstrated the effectiveness of employing gradient-based search across the feature dimension for PTS data in identifying an effective set of predictors for machine learning tasks. Our proposed algorithm, FSDR, consistently outperforms MI and LASSO. While the performance in terms of $R^2$ and $RMSE$ is comparable with that of SFS, we demonstrated that the execution time for SFS sharply increases with the increase in original feature size and target feature size, whereas the execution time for FSDR mainly depends only on the target size. Also, the influence of target size on the execution time is significantly lower in case of FSDR than that of SFS. Furthermore, despite being a DL-based algorithm, it can be effectively trained within only 871 data points to select features from a pool of as many as 4,200 original features.

As mentioned in Section \ref{sec-res}, a significant time required for FSCR operation is in generation and evaluation of the transformed continuous functions. It requires further investigations how the performance and time complexities are affected with simpler interpolation methods than Cubic Spline Interpolation, for example, Piecewise Linear Interpolation. 

%Furthermore, there is a high degree of multicollinearity in the  LUCAS dataset. Since FSDR can capture nonlinear relationships between predictors and the response variable unlike SFS and other commonly used FS algorithms, we propose a hypothesis that FSDR will significantly outperform SFS in datasets exhibiting a low degree of multicollinearity. In future studies, we plan to experiment with such datasets to validate this hypothesis.

%sensitive to initialization

\bibliographystyle{IEEEbib}
\bibliography{icme2023template}

\end{document}